\documentclass[journal]{IEEEtran}
\usepackage{graphicx} 
\usepackage{caption}
\usepackage{multirow}
\usepackage{amsmath}
\usepackage{amssymb}
\usepackage{algorithm}
\usepackage{algorithmic}
\usepackage{color}
\usepackage{booktabs}
\usepackage{multirow}
\usepackage{algorithm}
\usepackage{algorithmic}
\usepackage{color}
\usepackage{xcolor}
\usepackage{hyperref}
\ifCLASSINFOpdf
\else
\fi
\begin{document}
%
\title{Learning Disentangled Representation Implicitly via Transformer for Occluded Person Re-Identification}
%
%
%

\author{Mengxi~Jia,
        Xinhua~Cheng,
        Shijian~Lu
        and~Jian~Zhang
\IEEEcompsocitemizethanks{\IEEEcompsocthanksitem 
Manuscript received July 6, 2021. This work was supported in part by Key-Area Research and Development Program of Guangdong Province
(2019B121204008) and National Natural Science Foundation of China (61902009). \textit{(corresponding author: Jian Zhang.)}
\IEEEcompsocthanksitem  Mengxi Jia, Xinhua Cheng and Jian Zhang are with the School of Electronic and Computer Engineering, Shenzhen Graduate School, Peking University, Shenzhen 518055, China. (E-mail: \{mxjia, zhangjian.sz\}@pku.edu.cn)\protect
\IEEEcompsocthanksitem Shijian Lu is with the School of Computer Science and Engineering, Nanyang Technological University, Singapore 639798, Singapore. (E-mail: shijian.lu@ntu.edu.sg)
}
}

%
%

\markboth{Journal of \LaTeX\ Class Files}%
{Jia \MakeLowercase{\textit{et al.}}: Bare Demo of IEEEtran.cls for IEEE Journals}
%



\maketitle

\begin{abstract}
Person re-identification (re-ID) under various occlusions has been a long-standing challenge as person images with different types of occlusions often suffer from misalignment in image matching and ranking. Most existing methods tackle this challenge by aligning spatial features of body parts according to external semantic cues or feature similarities but this alignment approach is complicated and sensitive to noises. We design DRL-Net, a disentangled representation learning network that handles occluded re-ID without requiring strict person image alignment or any additional supervision. Leveraging transformer architectures, DRL-Net achieves alignment-free re-ID via global reasoning of local features of occluded person images. It measures image similarity by automatically disentangling the representation of undefined semantic components, e.g., human body parts or obstacles, under the guidance of semantic preference object queries in the transformer. In addition, we design a decorrelation constraint in the transformer decoder and impose it over object queries for better focus on different semantic components. To better eliminate interference from occlusions, we design a contrast feature learning technique (CFL) for better separation of occlusion features and discriminative ID features. Extensive experiments over occluded and holistic re-ID benchmarks (Occluded-DukeMTMC, Market1501 and DukeMTMC) show that the DRL-Net achieves superior re-ID performance consistently and outperforms the state-of-the-art by large margins for Occluded-DukeMTMC. Code is available at \href{github.com/Anonymous-release-code/DRL-Net}{\color{magenta}https://github.com/Anonymous-release-code/DRL-Net}.
\end{abstract}


%
\IEEEpeerreviewmaketitle

\begin{figure}[t]
\centering
\includegraphics[width=1.0\columnwidth]{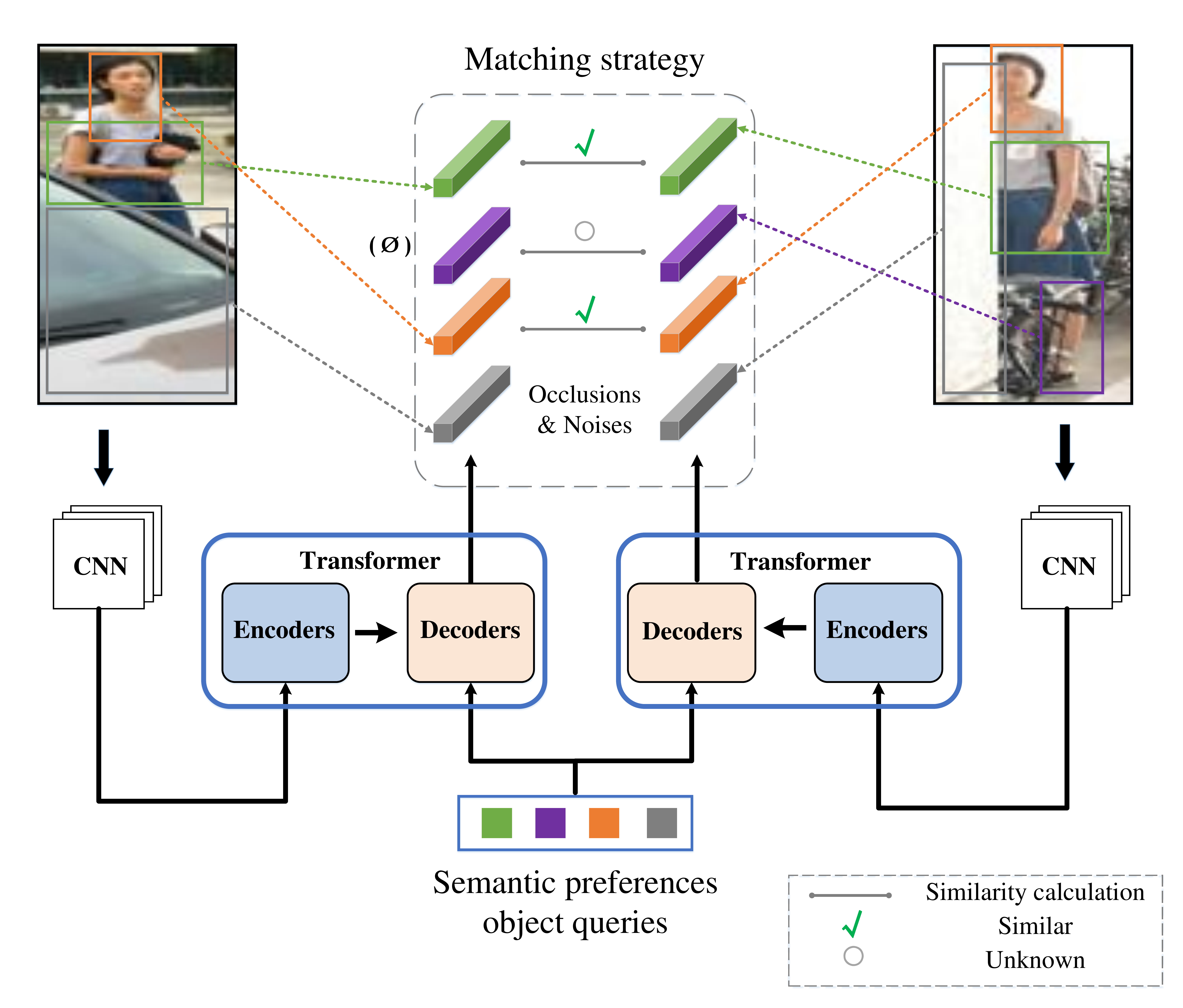}
\caption{
Illustration of the proposed DRL-Net in occluded person re-ID: Person images often suffer from various occlusions with different visible and invisible body parts which greatly complicate image alignment and similarity computation. The proposed DRL-Net is alignment-free which exploits semantic preferences object queries that guide transformer to disentangle the representation of body parts and eliminate occlusion noises in similarity measurement.}
\label{fig1}
\end{figure}

\section{Introduction}

\IEEEPARstart{P}ERSON re-identification (re-ID) \cite{zheng2016person} is a computer vision task that aims to associate person images captured by non-overlapping cameras. It has been studied intensively in recent years due to its wide applications in various video surveillance tasks \cite{gong2014re, zhou2017large, wang2019incremental, luo2019strong, cao2020progressive}. Thanks to the advance in deep learning and large-scale benchmarks, the re-ID research has achieved substantial progress and different approaches have been successfully proposed to tackle variations in viewpoints and poses \cite{wang2015zero}, illumination conditions \cite{zeng2020illumination}, camera configurations \cite{wang2014camera}, etc. On the other hand, most existing holistic re-ID methods \cite{wei2018glad} assume that the entire human body is visible in person images which hence cannot generalize well to occluded person images that suffer from incomplete information with various invisible body parts. Since humans are often occluded by clutters and obstacles in natural scenes, occluded re-ID \cite{ZhuoCLW18, Miao2019PoseGuidedFA} has great values in different surveillance tasks which is worth further investigation despite the complication resulting from missing body-part information.

Occluded re-ID is facing two major challenges. The first is super-rich variation of occlusions that block different body parts randomly and change the appearance of person images substantially. The occlusions thus introduce more intra-class variations which lead to more image matching errors and degraded re-ID performance. The second is interference of occlusions which often shares similar appearance as body parts and deteriorates the learnt person image representations. 
Most existing methods address the occlusion challenge by detecting the non-occluded body parts and aligning visible human parts in person image matching, and two typical alignment approaches have been widely investigated. The first approach exploits various external cues such as person masks \cite{Song2018MaskGuidedCA, He2019ForegroundAwarePR}, semantic parsing \cite{Kalayeh2018HumanSP} and pose estimation \cite{Miao2019PoseGuidedFA, GaoWLL20, Wang2020HighOrderIM} for accurate alignment of visible body parts.
However, the extraction of external cues is sensitive which tends to fail while facing severe occlusions and background noises. The other approach aligns body parts based on the similarity of local image features \cite{zheng2015partial, He2018DeepSF, Sun2019PerceiveWT, luo2020stnreid}, but it often struggles in differentiating human bodies from obstacles which often leads to mismatches. Beyond that, both alignment approaches involve complicated extra operations that take time in inference and also tend to accumulate errors.

This paper presents DRL-Net, an alignment-free re-ID framework that handles occlusions through disentangled representation learning as illustrated in Fig. \ref{fig1}. Leveraging the transformer architecture \cite{NIPS2017_3f5ee243}, DRL-Net eliminates the error-prone person alignment operations which first extracts compact image representations using CNNs and then performs global reasoning and ID prediction using the transformer encoder and decoder.

Specifically, DRL-Net disentangles the representations of undefined semantic components in occluded person images based on object queries without any additional supervision. Under the guidance of semantic preferences object queries, it adapts the transformer architecture that disentangles CNN features together with positional encoding into ID-relevant features (for person image matching) and ID-irrelevant features (for eliminating occlusion interference). DRL-Net performs global reasoning based on the interrelation of undefined semantic components, which allows feature disentanglement without any supervision of part correspondences and accordingly avoids the complicated and error-prone alignment process. For the transformer decoder, we impose a decorrelation constraint over semantic preference object queries to force them to focus on respective semantic components. In addition, we design a contrast feature learning module and a data augmentation strategy for better isolating ID-irrelevant features from global representation and suppressing occlusion interference.

The main contributions of this work are three-fold.
\begin{itemize}
    \item We propose a novel transformer framework DRL-Net that tackles occluded person re-ID by learning disentangled representation implicitly without any additional supervision and complicated alignment process. 
    
    \item We design a novel contrast feature learning technique together with a data augmentation strategy that mitigate the interference of occlusion noises effectively.
    
    \item The proposed DRL-Net achieves state-of-the-art re-ID performance under various occlusions yet without sacrificing performance over normal re-ID data with little occlusion. 

\end{itemize} 

\begin{figure*}[t]
\centering
\includegraphics[width=1.0\textwidth]{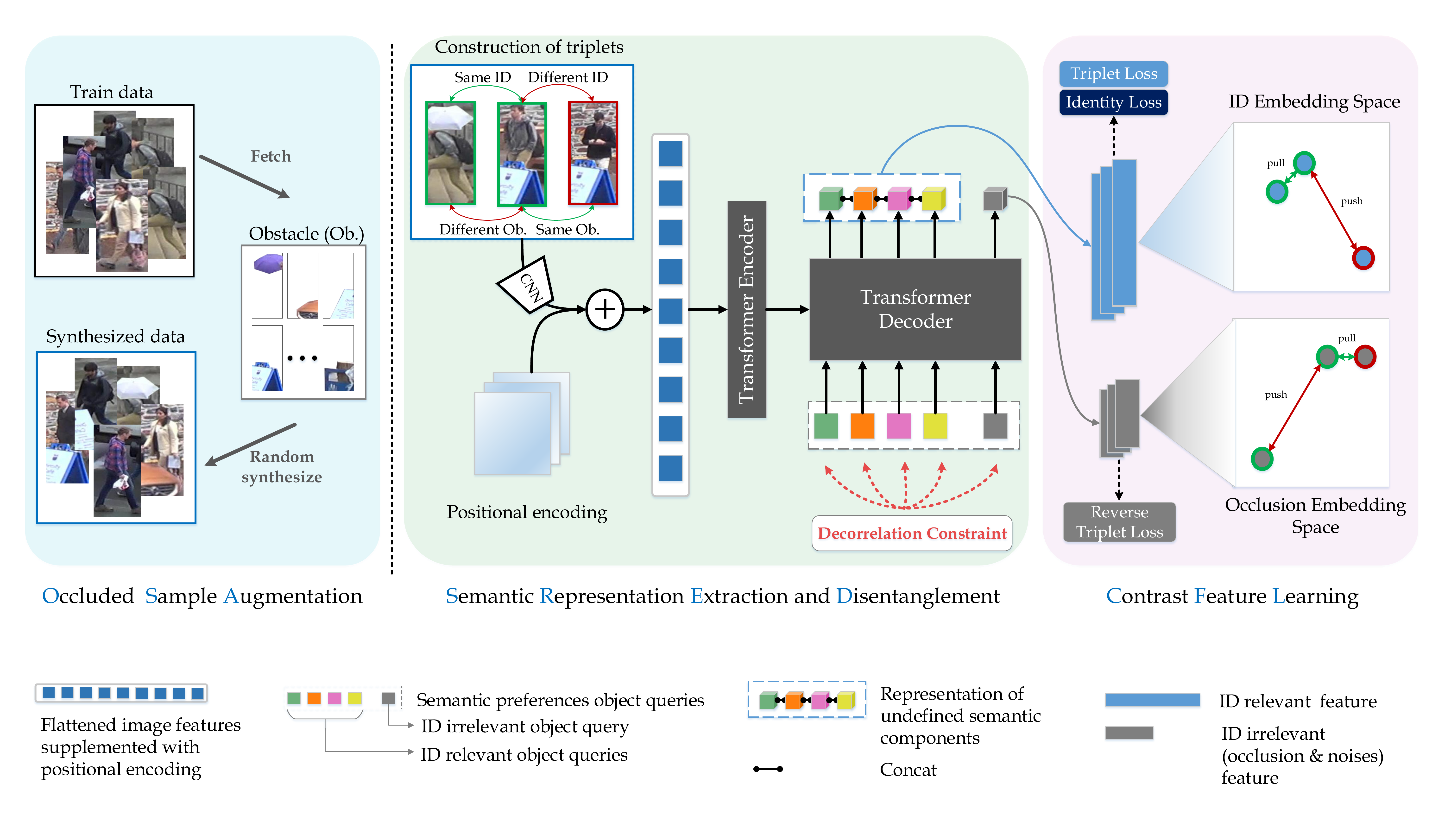} 
\caption{
The framework of the proposed DRL-Net: DRL-Net consists of three components. The first component is \textbf{occluded sample augmentation} that synthesizes person images by inserting various obstacles. The second component is \textbf{semantic representation extraction and disentanglement} that disentangles the representation of undefined semantic components into ID-relevant features and ID-irrelevant features under the guidance of semantic preference object queries. The third component is \textbf{contrast feature learning} that isolates ID-irrelevant features from global representation by optimizing features in both ID embedding space and occlusion embedding space reversely. Additionally, a decorrelation constraint is imposed over the object queries of the decoder to force them to focus on non-overlapped semantic parts.
}
\label{fig2}
\end{figure*}

\section{Related Work}
Person Re-ID has been one of the most studied problems due to its important application, and most of existing works were developed for matching holistic person that cannot tackle the occluded Re-ID problem. Since our method is proposed for occluded re-ID and based on transformer architecture, we only briefly review several related works in this section. 
\subsection{Occluded Person Re-ID}
The challenges of occluded re-ID mainly lie in body information incompleteness and spatial misalignment. Existing occluded re-ID methods can be roughly summarized into two streams, approaches with external cues and approaches based on part-to-part matching.

Previous methods leverage external cues such as human parsing, pose estimation or foreground segmentation to align parts of bodies. Under the guidance of extra semantic labels, such methods align parts precisely and benefit the feature representation. Miao et al. \cite{Miao2019PoseGuidedFA} propose a pose-guided feature alignment method (PGFA), taking advantage of the human semantic key-points to guide the matching of probe and gallery images. Gao et al. \cite{GaoWLL20} present a pose-guided visible part matching algorithm (PVPM) which jointly learns features and predicts the part visibility with attention heatmaps guided by pose estimation and graph matching accordingly. Wang et al. \cite{Wang2020HighOrderIM} propose a framework jointly modeling high-order relation and human-topology information by utilizing key-points estimation for robustly aligned features. However external cues requiring limits their usage and robustness in practical deployment. The inference of extra modules costs more time inevitably, and the generated semantic labels are untrustworthy under severe occlusions or low-resolution scenarios.

Models based on Part-to-part matching strategy handle occlusions by generating part alignment relations according to the similarity of local features across query and gallery images. Sun et al. \cite{sun2018beyond}
propose a network named Part-based Convolutional Baseline (PCB) which divided feature maps into horizontal pieces to learn local features directly. Zhang et al. \cite{Zhang2017AlignedReIDSH} align local features and compute distance by finding the shortest path. Zhu et al. \cite{Zhu2020IdentityGuidedHS} locate human body parts and potential person belongings at pixel-level by clustering algorithms to alignment. These methods match local features through self-supervision without external cues. Nevertheless, such auto-alignment steps require complicated algorithms like shortest path finding and clustering, and predict results strongly influenced by the way of dividing images.
Different from above strict alignment-based approaches, our method addresses the occluded person re-ID by leveraging transformer architectures, which can automatically and implicitly extract and disentangle the representations of target person without any additional supervision.
\subsection{Visual Transformer}
Transformer is a type of deep neural network which utilizes the self-attention mechanism and shows great performance on natural language processing tasks. Inspired by the significant success of transformer in the NLP field \cite{devlin-etal-2019-bert, pmlr-v80-oord18a, gu2018nonautoregressive, ghazvininejad-etal-2019-mask}, researchers applied transformer to various computer vision areas. Carion et al. \cite{Carion2020EndtoEndOD} present detection transformer (DETR) to view object detection as a direct set prediction problem, which firstly bring transformer architecture in high-level vision task. Vision transformer (ViT) proposed by Dosovitskiy et al. \cite{Dosovitskiy2020AnII} apply pure transformer and treats image patches as sequences directly, which achieved state-of-the-art performance on image recognition benchmarks. Now transformer are extended to more vision tasks including image processing \cite{Chen2020PreTrainedIP}, segmentation \cite{Xie2021SegmentingTO, Zheng2020RethinkingSS}, pose estimation \cite{Huang2020HandTransformerNS, Huang2020HOTNetNT}, etc.

We extend DETR to occluded re-id tasks, using object queries to extract features of semantic components instead of extra prediction or artificial pre-definition. Benefit from the outstanding representation capabilities of transformer architectures, our method achieves promising performance.

\section{Proposed Method}
In this section, we firstly introduce the architecture of the proposed DRL-Net in Section \ref{section 1}, which consists of a CNN and a Transformer. We then elaborate the designed contrast feature learning strategy for the DRL-Net in Section \ref{section 2}, which suppresses occlusion interference by separation of occlusion features and discriminative ID features. In Section \ref{section 3}, we explain the training and inference strategies in details. An overview of our method is shown in Fig. \ref{fig2}.


\subsection{Semantic Representation Extraction and Disentanglement}
\label{section 1}
\subsubsection{Feature Extractor}
Our feature extractor contains a CNN backbone and encoder-decoder layers, in order to extract compact representations and generate features of semantic component accordingly.

For a person image $x$, the CNN backbone generates feature maps $\textbf{f} = CNN(x)\in \mathbb{R}^{C\times H\times W}$, where $C, H, W$ denote the channel dimension, height and width of the feature maps respectively. With the non-linear activation function $\sigma(\cdot)$, we obtain the activated feature maps $\textbf{a} = \sigma(\textbf{f})\in \mathbb{R}^{C\times H\times W}$. A $1\times 1$ convolution layer is followed to generate the new feature maps $\textbf{g}\in \mathbb{R}^{d\times H\times W}$, where $d$ is smaller than $C$ to reduce the computation complexity of transformer. In order to construct the sequence form that transformers expect, we flatten the tensor along the last two spatial dimensions and finally get the $\textbf{g}\in \mathbb{R}^{d\times HW}$.

The encoder-decoder layers in our feature extractor follow the standard architecture of the transformer. We apply learnable positional encodings to encode spatial information and add it to the input of each encoder attention layer. To produce features of semantic components, we define semantic preferences object queries, which are a set of learnable input embeddings for decoder layers. More specifically, there are $N_q-1$ human semantic object queries and 1 occlusion object query, where $N_q$ is the semantic preferences object query number. 
The semantic preferences object queries denoted as $\textbf{Q}=[\textbf{q}_0, ..., \textbf{q}_{N_q-1}; \textbf{q}_o], \textbf{Q}\in \mathbb{R}^{N_q\times d}$ are different to generate representation features and they are added to the input of each attention layer. The representation features of undefined semantic components generated by $\textbf{Q}$ and encoder-decoder layers are denoted as $\textbf{F}=[\textbf{f}_0, ..., \textbf{f}_{N_q-1}; \textbf{f}_o], \textbf{F}\in \mathbb{R}^{N_q\times d}$. Features of human semantic components generated by $N_q-1$ ID relevant queries are concatenated as the ID relevant feature $\textbf{f}=concat([\textbf{f}_0, ...,  \textbf{f}_{N_q-1}])\in \mathbb{R}^{(N_q-1)\cdot d}$, and ID irrelevant feature $\overline{\textbf{f}}=\textbf{f}_o\in \mathbb{R}^{d}$ generated by occlusion query are used to reduce the interference of occlusions and noises.

We adopt cross entropy loss as identity loss to supervise the learning of feature extractor, and label smoothing is used to prevent the model from overfitting training IDs, which is defined as:

\begin{equation}
\begin{aligned}
&\mathcal{L}_{c e}=-\sum_{n=1}^{N} \sum_{m=1}^{M} q_{m} \log \mathcal{P}_{m}\left(\mathbf{f}_{n}\right), \\
&q_{m}=\left\{\begin{array}{ll}
1-\epsilon+\frac{\epsilon}{M} & \text { if } m=y_{n} \\
\frac{\epsilon}{M} & \text { otherwise,}
\end{array}\right.
\end{aligned}
\label{eq:ce}
\vspace{5pt}
\end{equation}
where $N$ is the number of training samples, $M$ is the person identity number of the training set, $\mathcal{P}_m(\textbf{f}_n)$ is the predicted probability of feature $\textbf{f}_n$ belonging to identity $m$, and $y_n$ is the ground-truth label of $\textbf{f}_n$. $q_m$ is the smoothing label according to the label $y_n$  and $\epsilon$ is a small constant and set to be $0.1$.

\subsubsection{Object Query Decorrelation Constraint}
To extract semantic-aligned features without external supervising, we expect features decoded from different object queries represent different semantic components. We propose object query decorrelation constraint to make object queries orthogonal with each other. Giving the set of object queries $\textbf{Q}^i\in \mathbb{R}^{N_q\times d}$ extracted from person image $i$, the object query decorrelation constraint loss is computed using the following formula:

\begin{equation}
\mathcal{L}_o=\alpha\sum^{N}_{i=1}\sum^{N_q}_{n=1}\sum^{N_q}_{m=1}abs({\frac{\left<\textbf{q}^{i}_n, \textbf{q}^{i}_m\right>}{||\textbf{q}^{i}_n||\ ||\textbf{q}^{i}_m||}}),
\label{eq:constraint}
\vspace{5pt}
\end{equation}
where $abs(\cdot)$ denotes the absolute value function, $\left<\cdot, \cdot\right>$ denotes the inner product, and $\alpha$ is the penalty factor of decorrelation constraint loss. 

The proposed decorrelation constraint is imposed over different object queries to force them to focus on respective semantic components with few overlaps, which helps the transformer better separate and localize the representation of different semantic components.  

\subsection{Semantic Preferences guided Contrast Feature Learning}
\label{section 2}
\subsubsection{Occluded Sample Augmentation (OSA)}
\label{OSA}

Occluded Sample Augmentation is a data augmentation strategy for our semantic preferences guided contrast feature learning. The limited number of occluded samples in training data often leads to the low diversity of occluded samples in each training batch, which makes the re-ID model sensitive to occlusions.
To address this issues, we employ OSA to augment person images which can preserve the person identities while generating new person images contains multiple obstacles. We first select different obstacles appearing in the train set as obstacle set $\mathcal{X}_{obstacle}$. During the training stage, we randomly selected $k$ obstacles from the $\mathcal{X}_{obstacle}$ to synthesize augmented samples for each training batch. Specifically, given an image batch $\mathcal{B}$ and random $k$ obstacles $[\textbf{o}_1, ..., \textbf{o}_k] \in \mathcal{X}_{obstacle}$, for each $\textbf{x}_i \in \mathcal{B}$ with label $y_{i}$, we generate augmented image $[\textbf{x}_{i,1}, ..., \textbf{x}_{i,k}]$ with label $y_{i}$ which occluded by $k$ obstacles. In this way,  the sample number in each batch increase by a factor of $k$. The augmented images together with original images are used for contrast feature learning (CFL). The benefits of employing OSA can be further demonstrated by introducing CFL.

\subsubsection{Contrast Feature Learning (CFL)}
\label{CFL}
We proposed semantic preferences guided contrast feature learning to expect semantic components generated by object queries to focus on body parts without the disturbance of occlusions. More specifically, we construct contrast triplets for given person image $\textbf{x}$ with the help of OSA, consisting of $\textbf{x}$ itself as the anchor, a positive instance with the same ID but different obstacles, and a negative one with different IDs but the same obstacle. The triplet loss with contrast triplets is defined by:

\begin{equation}
\mathcal{L}_{tri}=\sum^N_{n=1}[\delta+\mathcal{D}(\textbf{f}_n, \textbf{f}_{n+})-\mathcal{D}(\textbf{f}_n, \textbf{f}_{n-})]_+,
\label{eq:tri}
\vspace{5pt}
\end{equation}
where $\textbf{f}_n$ denotes the ID relevant features of image $\textbf{x}_n$, and $\textbf{f}_{n+}, \textbf{f}_{n-}$ denote the features belonging to the same or different person with $\textbf{f}_n$ respectively. $\mathcal{D}(\cdot,\cdot)$ is the distance function between features and $\delta$ is a margin parameter.

Furthermore, we proposed reverse triplet loss to make ID irrelevant features focus on occlusions or noises. We reverse the positive instances and negatives in contrast triplets to guide occlusion object query extract occlusion semantic components in images. The reverse triplet loss is defined by:

\begin{equation}
\mathcal{L}_{rtri}=\sum^N_{n=1}[\delta+\mathcal{D}(\overline{\textbf{f}}_n, \overline{\textbf{f}}_{n-})-\mathcal{D}(\overline{\textbf{f}}_n, \overline{\textbf{f}}_{n+})]_+,
\label{eq:rtri}
\vspace{5pt}
\end{equation}
where $\overline{\textbf{f}}_n$ denotes the ID irrelevant feature representations of image $\textbf{x}_n$, and $\overline{\textbf{f}}_{n+}, \overline{\textbf{f}}_{n-}$ denote the features belonging to the same or different person with $\overline{\textbf{f}}_n$ respectively. With the proposed decorrelation constraint and CFL, we force occluded semantic components only extracted by ID irrelevant object query, making human semantic components extracted by ID relevant queries free from occlusions.

\subsection{Training and Inference}
\label{section 3}
The training and inference process of the proposed DRL-Net is shown in Algorithm \ref{alg:algorithm}. Before the training stage, the obstacle set is constructed by obtaining obstacles from training images. In the occluded sample augmentation stage, given a mini-batch of images for training, we generate augmented samples with random obstacles to obtaining positive pairs and negatives. The entire feature extractor containing convolutional layers and encoder-decoder layers are trained together with the overall loss. The overall loss is therefore calculated as:

\begin{equation}
\mathcal{L} = \mathcal{L}_{ce} + \mathcal{L}_o + \mathcal{L}_{tri} + \lambda\mathcal{L}_{rtri},
\label{eq:final_loss}
\vspace{5pt}
\end{equation}
where $\lambda$ is the scale factor of reverse triplet loss, and the scale factors of others are set to be 1.

In the inference stage, query and gallery images are the input to feature extractor without augmentation, and we utilize ID relevant feature $\textbf{f}$ to compute the distance between query images and gallery images, ignoring the ID irrelevant feature $\overline{\textbf{f}}$.

\begin{algorithm}[tb]
\caption{Proposed DRL-Net}
\label{alg:algorithm}
\renewcommand{\algorithmicrequire}{\textbf{Input:}}
\renewcommand{\algorithmicensure}{\textbf{Output:}}
\begin{algorithmic}[1] 
\REQUIRE Training/query/gallery: $\mathcal{X}_{train}$, $\mathcal{X}_{query}$, $\mathcal{X}_{gallery}$
\ENSURE  Distance Matrix $\mathcal{D}$
\STATE \textit{$\%$Data preparation}
\STATE Obtain the obstacle set $\mathcal{X}_{obstacle}$ from the train set $\mathcal{X}_{train}$.
\STATE \textit{$\%$Training stage}
\STATE Initialize the CNN network parameters $\Theta$.
\FOR{each mini-batch $\mathcal{B} \subset \mathcal{X}_{train}$} 
\STATE Create set $\mathcal{B}'=\varnothing$
\STATE Random select obstacle $[\textbf{o}_1, ..., \textbf{o}_k] \in \mathcal{X}_{obstacle}$.
\FOR{each $\textbf{x}_i \in \mathcal{B}$}
\STATE Generate augmented sample $[\textbf{x}_{i,1}, ..., \textbf{x}_{i,k}]$ by $\textbf{x}_i$ and $[\textbf{o}_1, ..., \textbf{o}_k]$.
\STATE Adding $[\textbf{x}_i; \textbf{x}_{i,1}, ..., \textbf{x}_{i,k}]$ to set $\mathcal{B}'$.
\ENDFOR
\STATE Extract ID relevant feature $\textbf{f}_i$ and irrelevant feature $\overline{\textbf{f}}_i$ of each $\mathbf{x}_i \in \mathcal{B}'$ using DRL-Net.
\STATE Calculate $\mathcal{L}_{ce}, \mathcal{L}_o, \mathcal{L}_{tri}, \mathcal{L}_{rtri}$ by Eqs.(\ref{eq:ce}, \ref{eq:constraint}, \ref{eq:tri}, \ref{eq:rtri}).
\STATE Optimize CNN parameters $\Theta$ according to Eq.(\ref{eq:final_loss}).
\ENDFOR
\STATE \textit{$\%$Inference stage} 
\FOR{each $\textbf{x}_q \in \mathcal{X}_{query}$, $\textbf{x}_g \in \mathcal{X}_{gallery}$ }
\STATE Extract $\textbf{f}_q,\textbf{f}_g$ of $\textbf{x}_q,\textbf{x}_g$ respectively using DRL-Net.
\STATE Calculate $\mathcal{D}(\textbf{f}_q, \textbf{f}_g)$ by cosine distance metric.
\ENDFOR
\STATE \textbf{return} $\mathcal{D}$
\end{algorithmic}
\end{algorithm}

\section{Experiments}
\subsection{Datasets and Evaluation Metrics}
The experiments are conducted on three person ReID datasets, including one occluded re-ID dataset Occluded-DukeMTMC and two widely used Holistic re-ID datasets MSMT1 and Market-1501. 

\noindent\textbf{Occluded-DukeMTMC} \cite{Miao2019PoseGuidedFA} is a split of DukeMTMC-reID \cite{ZhengZY17} which keeps occluded images and removes some overlap images. It contains 15,618 training images, 17,661 gallery images, and 2,210 occluded query images, which is by far the largest occluded re-ID datasets. The experiments on this dataset follow the standard setting \cite{Miao2019PoseGuidedFA} and the training, query, and gallery sets contain 9\%, 100\%, and 10\% occluded images, respectively. \textbf{Market-1501} \cite{zheng2015scalable} consists of 32,668 images of 1,501 identities captured by 6 camera views. Following the standard setting \cite{zheng2015scalable}, the whole dataset is divided into a training set containing 12,936 images of 751 identities and a testing set containing 19,732 images of 750 identities. \textbf{DukeMTMC}
\cite{ZhengZY17} contains of 36,411 images of 1,812 persons from 8 cameras. 16,522 images of 702 persons are randomly selected from the dataset as the training set, and the remaining images are divided into the testing set containing 2,228 query images and 17,661 gallery images. The setting is same to \cite{ZhengZY17}. \textbf{MSMT17} contains 126,441 images of 4,101 IDs
captured from a 15-camera network. The training set has
32,621 images of 1,041 identities, and the testing
set has 93,820 images of 3,060 identities. During inference, 11,659 images are randomly selected as query images and the other 82,161 images are used as gallery images from the testing set \cite{Wei_2018_CVPR}.

\noindent\textbf{Evaluation metric} 
We adopt Cumulative Matching Characteristic (CMC) curve and mean average precision (mAP) for evaluations. All experiments are conducted in the single query mode and don’t use Re-Ranking to further refine the matching results.

\subsection{Implementation Details}
\noindent\textbf{Data preprocessing.} All person images are resized to $256\times128$ in both training and inference stages. The training images are augmented with random horizontal flipping, random cropping and random erasing \cite{zhong2020random} with a probability of 0.5. We construct an occlusion set by fetch obstacles which hardly appeared in test images from train set. All train images are copied and synthesized with random obstacles from our occlusion set in occluded sample augmentation stage.

\noindent\textbf{Backbones.} We adopt ResNet-50 \cite{HeZRS16} as the convolution neural backbone network. Following the setting of most ReID methods \cite{luo2019bag} , the last spatial down-sampling operation in ResNet-50 is removed to increase the spatial size of the feature map. In this case, the size of the feature map is $2048\times16\times8$. The hidden dimension $d$ is set to 256. The transformer layers are same with DETR and initialized with Xavier init \cite{glorot2010understanding}. The numbers of encoder layers, decoder layers and multi-head attention are set to 2, 2, 8 respectively. The cosine distance is used to measure the distance between probe and gallery images. 

\noindent\textbf{Optimization.} The CNN backbone network is pertrained over ImageNet \cite{krizhevsky2012imagenet}. Adam optimizer is adopted and we warm up the model for 10 epochs with a linearly growing learning rate from  $3.5\times10^{-5}$ to $3.5\times10^{-4}$. The learning rate is decreased by a factor of 0.1 at 40th and 70th epoch. The batch size is set to 32 with 4 images per ID. 

\begin{table}[t]
    \centering
    \caption{Comparison with state-of-the-art methods of occluded re-ID on Occluded-DukeMTMC. It shows that DRL-Net is superior to all three types of methods including the methods designed for holistic re-ID in the 1st group, utilizing external cues in the 2nd group and adopting part-to-part matching strategy in the 3rd group.}
    \label{tab:Occluded}
    \begin{tabular}{ l|cccc}
        \hline\hline
        Methods & Rank-1 & Rank-5 & Rank-10 & mAP \\
        \hline
        DIM (ArXiv 17) &21.5 &36.1 &42.8 & 14.4 \\
        Part Aligned (ICCV 17) & 28.8 &44.6 &51.0 & 20.2 \\
        HACNN (CVPR 18) & 34.4 &51.9 &59.4 & 26.0 \\
        Adver Occluded (CVPR 18) & 44.5 &- &- & 32.2 \\
        PCB (ECCV 18) & 42.6  &57.1 &62.9 & 33.7 \\
        \hline
        Part Bilinear (ECCV 18) & 36.9 &- &- & - \\
        FD-GAN (NIPS 18) & 40.8  &- &- & - \\
        PGFA (ICCV 19) & 51.4 &68.6 &74.9 & 37.3 \\
        HONet (CVPR 20) & 55.1 &- &- & 43.8 \\
        \hline
        DSR (CVPR 18) & 40.8 & 58.2 & 65.2 & 30.4 \\
        SFR (ArXiv 18) & 42.3 & 60.3 & 67.3 & 32.0 \\
        MoS (AAAI 21) & 61.0 & 74.4 & 79.1 & 49.2 \\
        \hline
        \textbf{DRL-Net} $(Ours)$ & \textbf{65.0} & \textbf{79.3} & \textbf{83.6} & \textbf{50.8}\\ 
        \hline\hline
    \end{tabular}
\end{table}

\begin{table}[t]
    \centering
    \caption{Ablation study over Occluded-DukeMTMC. T, OSA and CFL denotes proposed transformer architecture, occluded sample augmentation and contrast feature learning respectively.}
    \label{tab:Ablation}
    \begin{tabular}{ l|cccc}
        \hline\hline
        Methods & Rank-1 & Rank-5 & Rank-10 & mAP \\
        \hline
        Baseline & 51.0 & 66.3 & 71.7 & 43.8 \\
        Baseline+T & 59.6 & 74.4 & 79.5 & 49.0 \\
        Baseline+T+OSA & 60.5 & 74.7 & 80.9 & 48.2 \\
        Baseline+T+OSA+CFL & 65.0 & 79.3 & 83.6 & 50.8 \\
        \hline\hline
    \end{tabular}
\end{table}

\subsection{Comparison with the State-of-the-Art}
\begin{table}[t]
    \centering
    \caption{Comparison over datasets Market-1501 and DukeMTMC shows DRL-Net can be generalized to holistic re-ID with superior performance: The compared methods are grouped into three categories: global feature based, part feature based and external cues based.}
    \label{tab:comparison}
    \begin{tabular}{l|c c|c c}
        \hline\hline
        \multirow{2}{*}{Methods} & \multicolumn{2}{c|}{Market-1501} & \multicolumn{2}{c}{DukeMTMC} \\
        \cline{2-5}
         & Rank-1 & mAP & Rank-1 & mAP \\
        \hline
        IANet (CVPR 19) & 94.4 & 83.1 & 87.1 & 73.4 \\
        MVPM (ICCV 19) & 91.4 & 80.5 & 83.4 & 70.0 \\
        DMML (ICCV 19) & 93.5 & 81.6 & 85.9 & 73.7 \\
        SFT (ICCV 19) & 93.4 & 82.7 & 86.9 & 73.2 \\
        VCFL (ICCV 19) & 89.3 & 74.5 & - & - \\
        Circle (CVPR 20) & 94.2 & 84.9 & - & - \\
        \hline
        PCB(ECCV 18) & 92.3 & 77.4 & 81.8 & 66.1 \\
        PCB+RPP (ECCV 18) & 93.8 & 81.6 & 83.3 & 69.2\\
        AlignedReID(ArXiv18) & 91.8 & 79.3 & - & - \\
        DSR (CVPR 18) & 83.6 & 64.3 & - & - \\
        VPM (CVPR 19) & 93.0 & 80.8 & 83.6 & 72.6 \\
        \hline
        SPReID (CVPR 18) & 92.5 & 81.3 & - & - \\
        MGCAM (CVPR 18) & 83.8 & 74.3 & 46.7 & 46.0 \\
        Pose-transfer (CVPR18) & 87.7 & 68.9 & 30.1 & 28.2 \\
        PSE (CVPR 18) & 87.7 & 69.0 & 27.3 & 30.2 \\
        PGFA (ICCV 19) & 91.2 & 76.8 & 82.6 & 65.5 \\
        AANet (CVPR 19) & 93.9 & 82.5 & 86.4 & 72.6 \\
        HONet (CVPR 20) & 94.2 & 84.9 & 86.9 & 75.6 \\
        \hline
        \textbf{DRL-Net} \emph{(Ours)} & \textbf{94.7} & \textbf{86.9} & \textbf{88.1} & \textbf{76.6} \\
        \hline\hline
    \end{tabular}
\end{table}

\begin{table}[t]
    \centering
    \caption{Comparison over holistic re-ID dataset MSMT17.}
    \label{tab:msmt17}
    \begin{tabular}{ l|cccc}
        \hline\hline
        Methods & Rank-1 & Rank-5 & Rank-10 & mAP \\
        \hline
        MVPM (ICCV 19) &71.3 &84.7 &- & 46.3 \\
        SFT (ICCV 19) &73.6 &85.6 &- & 47.6 \\
        DG-Net (CVPR 19) & 77.2 &87.4 &90.5 & 52.3 \\
        IANet (CVPR 19) & 75.5 &85.5 &88.7 & 46.8 \\
        Circle (CVPR 20) & 76.3 &- &- & 50.2 \\
        Circle + MGN (CVPR 20) & 76.9 &- &- & 52.1 \\
        \hline
        \textbf{DRL-Net} $(Ours)$ & \textbf{78.4} & \textbf{88.2} & \textbf{91.3} & \textbf{55.3}\\ 
        \hline\hline
    \end{tabular}
\end{table}

\begin{table}[t]
    \centering
    \caption{The comparison and analysis for the number $N_l$ of transformer layers over Occluded-DukeMTMC. The number of layers for encoder and decoder are the same.}
    \label{tab:layer_num}
    \begin{tabular}{ l|cccc}
        \hline\hline
        $N_l$ & Rank-1 & Rank-5 & Rank-10 & mAP \\
        \hline
        1 & 55.3 & 71.8 & 78.0 & 47.0 \\
        2 & 57.2 & 72.4 & 78.2 & 47.0 \\
        3 & 55.9 & 71.7 & 77.4 & 46.0 \\
        4 & 55.3 & 71.3 & 77.6 & 45.2 \\
        5 & 55.7 & 73.2 & 78.8 & 44.9 \\
        6 & 53.9 & 70.1 & 75.7 & 42.9 \\
        \hline\hline
    \end{tabular}
\end{table}

\begin{table}[t]
    \centering
    \caption{Parameter analysis of the number $N_q$ of object queries over Occluded-DukeMTMC. The results shows DRL-Net is robust to different $N_q$.}
    \label{tab:query_num}
    \begin{tabular}{ l|cccc}
        \hline\hline
        $N_q$ & Rank-1 & Rank-5 & Rank-10 & mAP \\
        \hline
        2 & 53.5 & 69.2 & 75.3 & 43.6 \\
        5 & 55.3 & 71.9 & 77.8 & 46.4 \\
        9 & 57.1 & 72.4 & 78.2 & 47.3 \\
        13 & 57.2 & 73.9 & 79.0 & 47.3 \\
        17 & 57.7 & 75.2 & 79.7 & 47.2 \\
        \hline\hline
    \end{tabular}
\end{table}

We compare our method with state-of-the-art methods for both occluded and holistic person re-ID tasks in Table \ref{tab:Occluded} and Table \ref{tab:comparison}, respectively. The backbones of compared methods  are ResNet-50 or modified ResNet-50 by using branches, attentions or different convolution operations.

\noindent\textbf{Results on Occluded-DukeMTMC.}
The comparision results over dataset Occluded-DukeMTMC are shown in Table \ref{tab:Occluded}. There are three mainstream types of occluded re-ID methods are compared: holistic re-ID methods without designing modules for occlusions (DIM \cite{Yu2017TheDI}, Part Aligned \cite{Zhao2017DeeplyLearnedPR}, HACNN \cite{Li2018HarmoniousAN}, Adver Occluded \cite{Huang2018AdversariallyOS} and PCB \cite{sun2018beyond}), Occluded re-ID methods with external cues (Part Bilinear \cite{suh2018part}, FD-GAN \cite{ge2018fd}, PGFA \cite{Miao2019PoseGuidedFA} and HONet \cite{Wang2020HighOrderIM}), and Occluded re-ID methods based on part-to-part matching (DSR \cite{He2018DeepSF}, SFR \cite{He2018RecognizingPB} and MoS \cite{jia2021matching}). Our DRL-Net method achieves 65.0\% Rank-1 accuracy, 79.3\% Rank-5 accuracy, 83.6\% Rank-10 accuracy, and 50.8\% mAP, which outperforming all types of methods by a large margin. 

The superior performance of DRL-Net is summarized into three aspects. First, the transformer layers enhance the representation capacity of CNN backbones. Second, the introduction of object queries and decorrelation constraint gives our method the ability to learn disentangled representation implicitly without external cues. Third, the novel metric learning CFL with synthesized images efficiently weakens the interference of occlusions and noises.

\noindent\textbf{Results on Market-1501 and DukeMTMC.}
The comparision results over holistic re-ID datasets including Market-1501 and Duke-MTMC are shown in Table \ref{tab:comparison}. Three types of holistic re-ID methods are considered for comparison: methods based on global features (IANet \cite{Hou2019InteractionAndAggregationNF}, MVPM \cite{SunCYX19}, DMML \cite{Chen2019DeepMM}, SFT \cite{Luo2019SpectralFT}, VCFL \cite{Liu2019ViewCF}, Circle \cite{sun2020circle}, MoS \cite{jia2021matching}), methods using part features (PCB, PCB+RPP \cite{sun2018beyond}, AlignedReID \cite{Zhang2017AlignedReIDSH}, DSR \cite{He2018DeepSF} and VPM \cite{Sun2019PerceiveWT}), and methods using external cues including human-parsing based (SPReID \cite{Kalayeh2018HumanSP} and MGCAM \cite{Song2018MaskGuidedCA}), attribute information based (AANet \cite{Tay2019AANetAA}) and human pose based (Pose-transfer \cite{Liu2018PoseTP}, PSE \cite{Sarfraz2018APE}, PGFA \cite{Miao2019PoseGuidedFA} and HONet \cite{Wang2020HighOrderIM}). Though DRL-Net is not proposed for the holistic re-ID task, it performs comparable results with all types of holistic re-ID methods, showing the robustness of our proposed methods.

\noindent\textbf{Results on MSMT17.}
Since MSMT17 is released recently, hence there are only a few methods that report on this dataset, including MVMP \cite{SunCYX19}, SFT \cite{Luo2019SpectralFT}, DG-Net \cite{Zheng_2019_CVPR}, IANet \cite{Hou2019InteractionAndAggregationNF}, Circle \cite{sun2020circle} and Circle + MGN \cite{sun2020circle}. Table \ref{tab:msmt17} shows the comparison results. DRL-Net achieves outstanding performance in all evaluation metrics.
\subsection{Ablation Study}
In this section, we conducted extensive ablation studies to investigate the effectiveness of each component of DRL-Net. We used ResNet-50 as backbone and performed ablation experiments over Occluded-DukeMTMC. Table \ref{tab:Ablation} shows experimental results. 

\noindent\textbf{Effectiveness of the proposed Transformer Architecture.}
We first study the effect of proposed transformer-based feature extractor which is denoted as \emph{baseline+T} by removing the CFL and OSA in the framework. The \emph{Baseline} model which directly uses ResNet-50 as feature extractor and the \emph{baseline+T} model are both trained by original triplet loss as well as identity loss. As shown in the first two rows of Table \ref{tab:Ablation}, consistent improvements are achieved on all four evaluation metrics. This indicates that the transformer has a strong ability for feature extracting and disentangling through conducting the global reasoning to further combine the features, which helps to handle the occlusion challenge effectively. The benefits of employing transformer architecture can be further demonstrated by introducing other relevant designs and operations.        

\noindent\textbf{Effectiveness of the proposed OSA.} We evaluate the occluded sample augmentation as described in Section \ref{OSA}. For this experiment, we design a network \emph{baseline+T+OSA} that just incorporates the occluded sample augmentation into the \emph{baseline+T} and maintain the training strategy. As shown in Table \ref{tab:Ablation}, occluded sample augmentation can improve the re-ID performance on CMC Rank-1/5/10. The improvement can be explained by the effectiveness of the augmented samples that increases the diversity of occluded training samples. On the other hand, due to the gap between the synthesized occluded images and the real images, it suffers a slight decrease in mAP when simply incorporating the occluded sample augmentation.

\noindent\textbf{Effectiveness of the proposed CFL.} We further evaluate the contrast feature learning component as described in Section \ref{OSA}. For this experiment, We incorporate contrast feature learning into the \emph{baseline+T+OSA} as described in the previous subsection and we denote it as \emph{baseline+T+OSA+CFL}. As shown in Table \ref{tab:Ablation}, the incorporation of contrast feature learning significantly improves the person re-ID performance beyond \emph{baseline+T+OSA}. The \emph{baseline+T+OSA+CFL} achieves a rank-1 accuracy of 65.0\% and an mAP of 50.8\% Which outperforms the corresponding \emph{baseline+T+OSA} by 4.5\% and 2.6\%, respectively. The effectiveness of the contrast feature learning can be largely attributed to the separation of occlusions feature and discriminative ID-relevant features, which is crucial to eliminate interference from occlusions for occluded person re-ID.

The ablation studies show that the proposed DRL-Net outperforms the \emph{Baseline} by 14.0\% in Rank-1 accuracy and 7.0\% in mAP while working with the occluded sample augmentation and contrast feature learning. This demonstrates that the three components complement each other in achieving better occluded re-ID performance.

\begin{figure}[t]
\centering
\includegraphics[width=0.95\columnwidth]{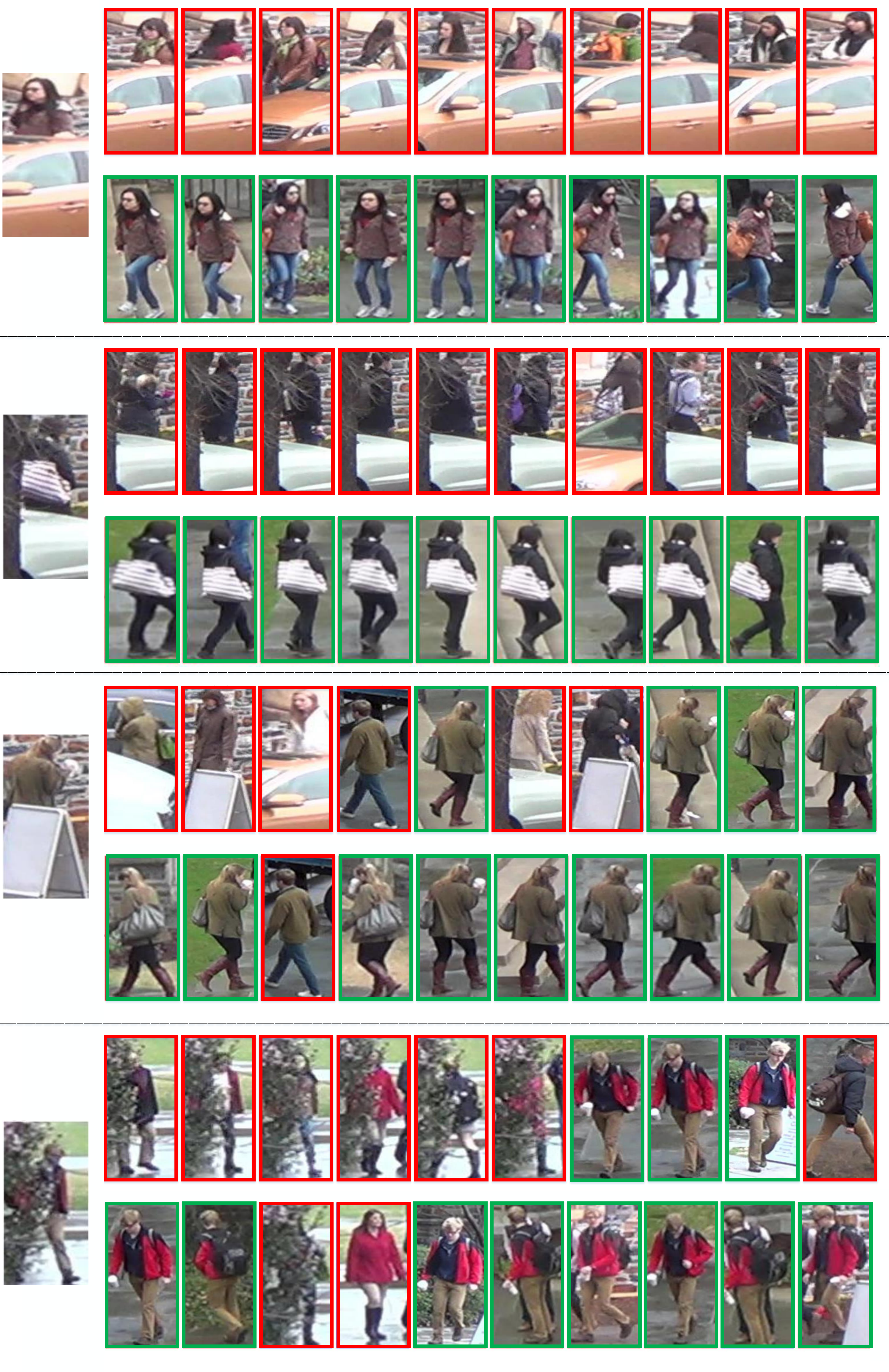} 
\caption{Illustration of the proposed DRL-Net: For each occluded query person image on the left, the two rows on the right show the ten top-matching images as returned by the \emph{baseline} model (in the first row) and the proposed DRL-Net (in the second row). The green and red boxes highlight positive and negative matching, respectively.}
\label{fig:ranklist}
\end{figure}

\begin{figure}[t]
\centering
\includegraphics[width=1.0\columnwidth]{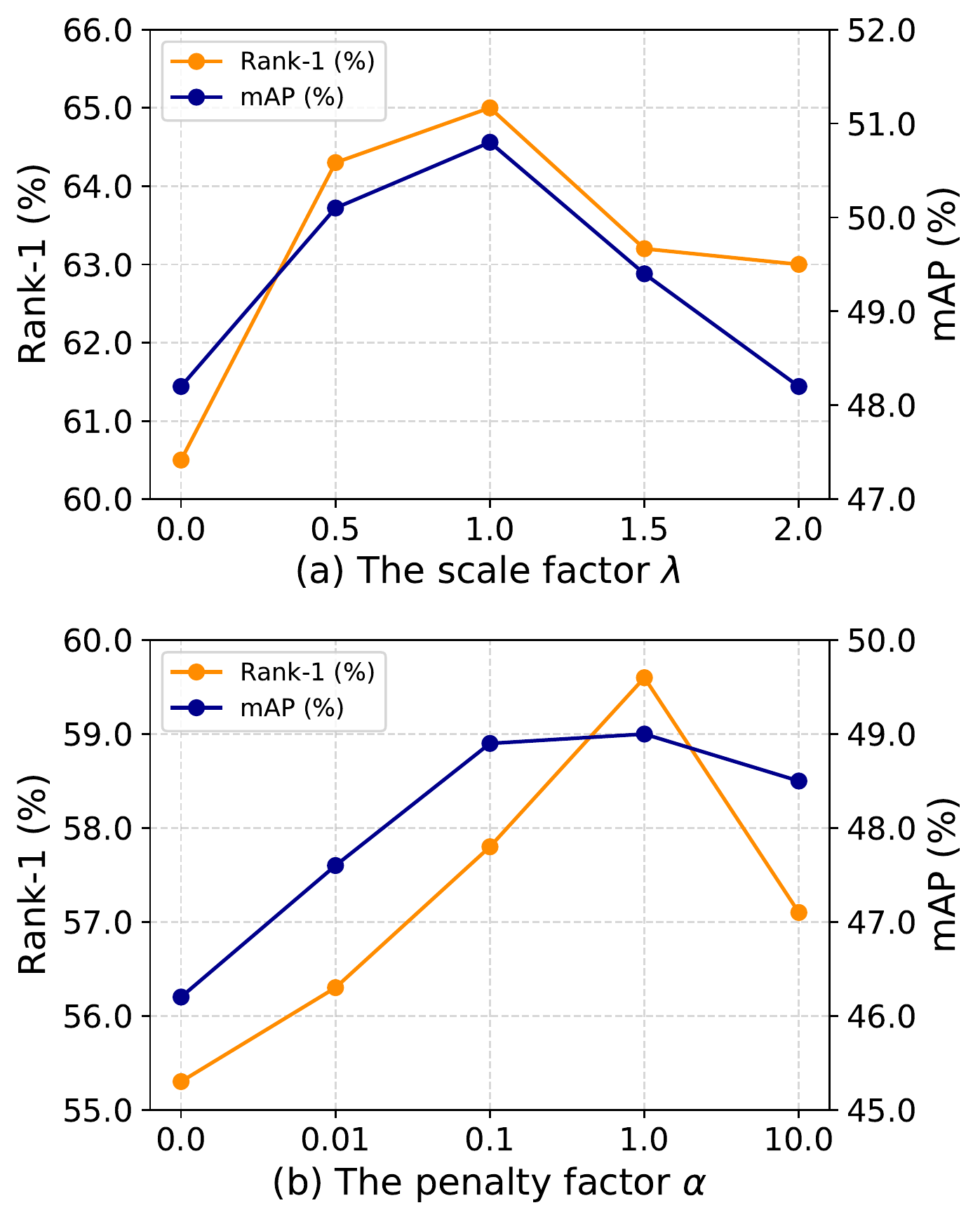}
\caption{Parameter analysis for the scale factor $\lambda$ and the penalty factor $\alpha$.
}
\label{fig3}
\end{figure}

\subsection{Parameter Analysis}
\noindent\textbf{The Number of Transformer Layers.} 
The impressive performance that transformer achieved can largely contribute to its self-attention mechanism, with which transformer can globally model relations between feature representations of different semantic components. To evaluate the importance of self-attention mechanism, we conduct experiments by changing the number of encoder-decoder layers $N_l$ as shown in Table \ref{tab:layer_num}. We observe that when the $N_l$ is set to 2, the best re-ID performance is achieved, and then the improvement brought by transformer diminishes as depth increases. We think it is because the re-ID task utilizes the lower-resolution representations throughout the network than other high-level vision tasks (e.g. object detection). Moreover, the scale of the re-ID datasets is relatively small and the image contents in datasets are simple, which makes the cross-correlations between the output elements of the decoder are easy to compute.


\noindent\textbf{The Number of Semantic Preferences Object Queries.}
Intuitively, the number of semantic preferences object queries $N_q$ determines the granularity of the semantic components. We perform the quantitative ablation studies to find the most suitable $N_q$. As detailed in Table \ref{tab:query_num}, the performance of DRL-Net is robust to different $N_q$. We can observe that as the $N_q$ increases, the re-ID performance is continuously improved, but the inference cost also increases correspondingly. To balance performance and cost, we set $N_q$ to 9 in the final version.

\noindent\textbf{The Robustness of Parameters $\lambda$ and $\alpha$.}
We studied hyper parameters in DRL-Net by setting it to different values and checking the person Re-ID performance. Fig \ref{fig3} shows the experimental results on Occluded-DukeMTMC dataset. We first analyze the influence of $\lambda$ in  Fig \ref{fig3} (a), the scale factor $\lambda$ in Eq. \ref{eq:final_loss} is the balancing weight of contrast feature learning (CFL). With $\lambda$ increasing, the Rank-1/mAP is improved by 4.5\%/2.6\% ($\lambda$ = 1.0), which means the CFL module now is beneficial for learning better occlusion robust re-ID features. Continuing to increase $\lambda$, the performance is degraded because the weights for human parts feature embedding and the position embedding are weakened.

Then we analyze the effect of penalty factor $\alpha$ on \emph{baseline+T} model as shown in Fig \ref{fig3} (b). The penalty factor $\alpha$ in Eq. \ref{eq:constraint} will affect correlation among object queries in decoder. Experiments show that \emph{baseline+T} performs best when $\alpha$ = 1.0. Using a smaller $\alpha$ will suppress the value of $\mathcal{L}_o$, lower the decorrelation ability for object queries. On the other hand, $\alpha$ should not be very large for preserving the feature representation capability of the model.
Experimental results both in (a) and (b) show that DRL-Net performs stably and is tolerant to the change of the parameters.


\subsection{Visualization}
\noindent\textbf{Qualitative Results.} 
We demonstrate how DRL-Net overcomes the occlusion constraint by providing several samples of person image ranking and Fig \ref{fig:ranklist} shows experimental results. For each occluded query person image on the left, the two rows of images on the right show the 10 top-matching images that are produced by the \emph{baseline} and our proposed DRL-Net, respectively. We can observe that DRL-Net can overcome the occlusions and identify images of the same pedestrian correctly (highlighted by green-color boxes). As a comparison, the \emph{baseline} network is very sensitive to occlusions obviously and returns a large amount of false-matching person images (highlighted by red-color boxes).

\begin{figure}[t]
\centering
\includegraphics[width=0.95\columnwidth]{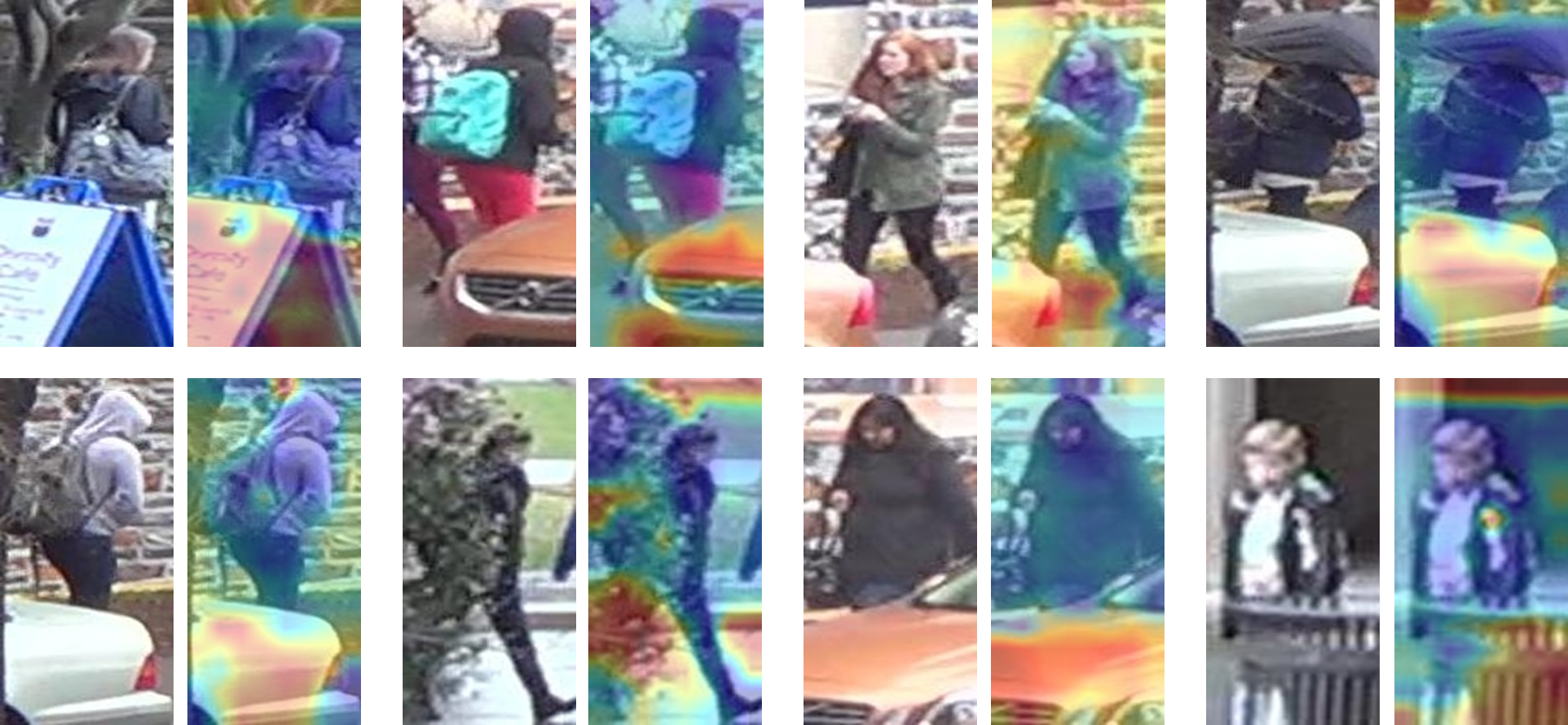} 
\caption{
Visualization of the decoder attention of ID-irrelevant object queries: For each of the eight image pairs, the left shows the original person image and the right shows the heat map of ID irrelevant object query.}
\label{fig:visual}
\end{figure}

\noindent\textbf{Visualizing for object query.} 
We visualize decoder cross-attentions for the ID-irrelevant object query (the last object query on decoder) using the heat map. The redder the heat map, the higher the attention scores. As Fig. \ref{fig:visual} shows, the ID-irrelevant object query can automatically localize occlusion and noise areas without explicit supervision, though obstacles various in different person images. This nice property is largely attributed to the transformer architecture and the contrast feature learning that guides the network to disentangle the representation of different semantic components and encourage the separation of occlusions feature and discriminative ID-relevant features.
\section{Conclusion}
This paper proposes a novel alignment-free method DRL-Net that handles occluded re-ID through disentangled representation learning. Leveraging transformer architectures, DRL-Net performs global reasoning based on the interrelation of undefined semantic components, which allows feature disentanglement without any supervision of part correspondences. Furthermore, to better eliminate the interference of occlusion noises, we design a contrast feature learning technique to encourage the separation of occlusions feature and ID-relevant features. Extensive experimental evaluations on several benchmarks demonstrate that DRL-Net achieves superior re-ID performance consistently.


\bibliographystyle{IEEEtran}
\bibliography{egbib}


%





\ifCLASSOPTIONcaptionsoff
  \newpage
\fi

\end{document}